\documentclass[10pt,twocolumn,letterpaper]{article}
\usepackage[pagenumbers]{cvpr} 

\usepackage{graphicx}
\usepackage{amsmath}
\usepackage{amssymb}
\usepackage{booktabs}

\usepackage{color}
\usepackage{multirow}
\usepackage{float}
\usepackage{booktabs}
\usepackage{comment}
\usepackage{colortbl}
\usepackage{tabularx}
\usepackage[labelfont=bf]{caption}
\usepackage{collcell}

\newcommand{\bb}[1]{\textbf{#1}}

\usepackage[pagebackref,breaklinks,colorlinks]{hyperref}

\usepackage[capitalize]{cleveref}
\crefname{section}{Sec.}{Secs.}
\Crefname{section}{Section}{Sections}
\Crefname{table}{Table}{Tables}
\crefname{table}{Tab.}{Tabs.}

\newcolumntype{K}[1]{>{\centering\arraybackslash}p{#1}}
\newcolumntype{S}[0]{>{\centering\let\newline\\\arraybackslash\hspace{0pt}}m{0.2cm}}
\newcolumntype{M}[0]{>{\centering\let\newline\\\arraybackslash\hspace{0pt}}m{0.4cm}}
\newcolumntype{Q}[0]{>{\centering\let\newline\\\arraybackslash\hspace{0pt}}m{1.0cm}}
\newcolumntype{C}[0]{>{\centering\let\newline\\\arraybackslash\hspace{0pt}}m{1.8cm}}

\begin{document}

\title{Audio-Visual Person-of-Interest DeepFake Detection}
\author{Davide Cozzolino\textsuperscript{1} \ \ \ 
Alessandro Pianese\textsuperscript{1} \ \ \ 
Matthias Nie\ss ner\textsuperscript{2} \ \ \ 
Luisa Verdoliva\textsuperscript{1} \\[2mm]
{\textsuperscript{1}University Federico II of Naples \ \ \ \ \ \textsuperscript{2}Technical University of Munich}}

\newcommand{\ru}{\rule{0mm}{3mm}}
\newcommand{\rota}[1]{\rotatebox[origin=c]{90}{#1}}
\newcommand{\NAME}{{POI-Forensics}}

\twocolumn[{%
	\renewcommand\twocolumn[1][]{#1}%
	\maketitle
	\begin{center}
	    \vspace{0.2cm}
		\includegraphics[page=2, width=0.8\linewidth, trim=0 132 0 30]{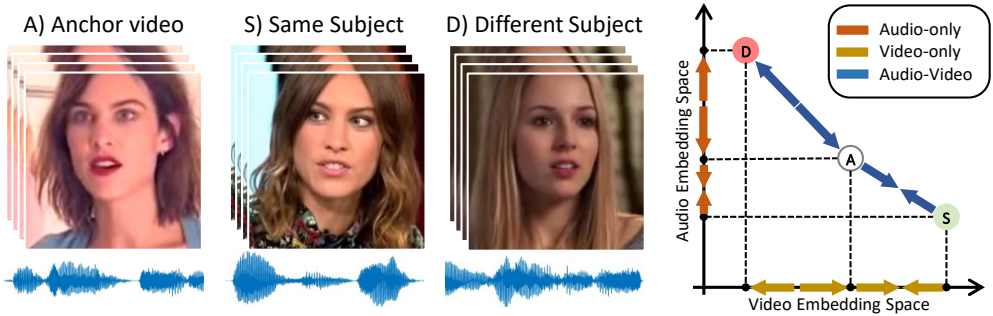}
	    \captionof{figure}{We propose \NAME, a method to detect Deepfakes based on audio-visual identity verification.
    Given a set of real identities, we propose a contrastive learning method that encourages embedded vectors of a reference video (A) to be close to embedded vectors of the same subject (S), but far from those of different subjects (D).}
		\label{fig:teaser}
	\end{center}
}]

\begin{abstract}
\vspace{-0.5cm}
Face manipulation technology is advancing very rapidly,
and new methods are being proposed day by day.
The aim of this work is to propose a deepfake detector that can cope with the wide variety of manipulation methods and scenarios encountered in the real world.
Our key insight is that each person has specific characteristics that a synthetic generator likely cannot reproduce.
Accordingly, we extract audio-visual features which characterize the identity of a person,
and use them to create a person-of-interest (POI) deepfake detector.
We leverage a contrastive learning paradigm to learn the moving-face and audio segment embeddings
that are most discriminative for each identity.
As a result, when the video and/or audio of a person is manipulated, 
its representation in the embedding space becomes inconsistent with the real identity, allowing reliable detection.
Training is carried out exclusively on {\em real} talking-face video; thus,
the detector does not depend on any specific manipulation method and yields the highest generalization ability.
In addition, our method can detect both single-modality (audio-only, video-only) and multi-modality (audio-video) attacks, and is robust to low-quality or corrupted videos.
Experiments on a wide variety of datasets confirm that our method ensures a SOTA performance, 
especially on low quality videos. Code is publicly available on-line at
\url{https://github.com/grip-unina/poi-forensics}.
\end{abstract}

\section{Introduction}

\begin{figure}[t!]
    \centering
    \includegraphics[page=8, width=\linewidth, clip, trim=0 145 0 0]{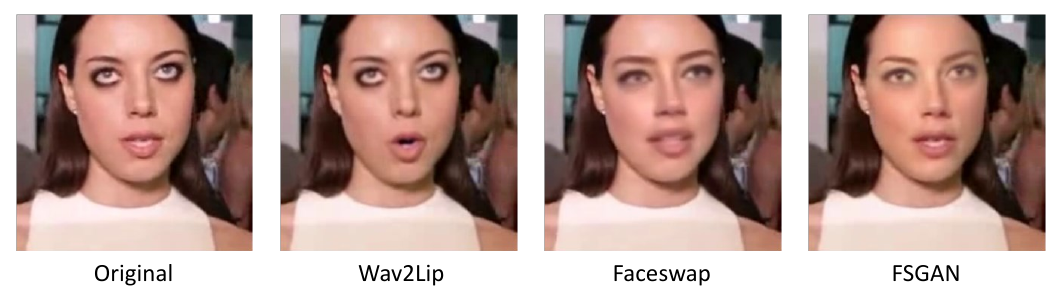}
    \caption{Different manipulations applied to the same original video, from left to right:
    original video, facial reenactment manipulation using Wav2Lip~\cite{Prajwal2020lip} and two faces-swapping-manipulation using Faceswap~\cite{Korshunova2017fast} and Faceswap GAN (FSGAN)~\cite{Nirkin2019fsgan}.}
    \label{fig:examples}
\end{figure}

Synthetic media generation has become a key technology in many industrial applications, from film production to the video game industry. Facial manipulations, however, also pose a serious and growing threat to our society, of which financial fraud and disinformation campaigns are just a few examples. 
With the advancement of such technology, there is a steady increase in the level of photorealism, as more and more methods of video manipulation emerge.
In particular, the term deepfake, which is often associated with face-swapping, has now become associated with negative implications.
Currently, the deepfake term has taken on an even broader meaning, including a variety of possible video manipulations: speech can be synthesized in anyone's voice, face expression can be modified, the identity of a person can be swapped with another, even altering what they are saying. Some examples of different manipulations of the same identity are shown in Fig.~\ref{fig:examples}: a video where the audio has been manipulated and lip movements have been perfectly synchronized with it \cite{Prajwal2020lip} and two different types of face-swapping \cite{Korshunova2017fast,Nirkin2019fsgan}.

Dealing with such a large spectrum of manipulations is the main challenge for current video deepfake detectors. In fact, it is particularly difficult to develop a method that can detect multiple known manipulation methods at the same time; this is only exacerbated when targeting unknown methods that were not part of any training samples. As a result, current SOTA detectors, trained over large datasets of deepfake and pristine videos, often show an unsatisfactory cross-dataset performance, clearly highlighting the limitations of the supervised deep learning based approach. In addition, performance often drops dramatically under a more challenging scenario, such as low-quality videos. Such conditions, however, are commonplace in the real world where videos are mostly disseminated through social networks where they are further compressed with the loss of relevant audio-visual information. It is also worth pointing out that deep learning-based methods are vulnerable to adversarial attacks with detection performance that degrades sharply even in a black-box scenario \cite{Hussain2021adversarial,Neekhara2021adversarial,Cozzolino2021spoc}.

A possible solution to gain generalization and robustness is to shift to a completely different paradigm, training models only on real videos, with the goal to detect manipulated videos based on their anomalous behavior \cite{Cozzolino2019extracting}. This approach turns out to be particularly effective if the characterization of pristine faces is based on semantic features, such as soft biometrics, leading to Person-of-Interest (POI) based detection \cite{Agarwal2019,Agarwal2020,Cozzolino2021idreveal, Agarwal2021watch, bohavcek2022protecting, dong2022protecting}. First papers on this topic exploited specific face and head movement patterns \cite{Agarwal2019,Agarwal2020, Cozzolino2021idreveal}, inconsistencies between mouth shape dynamics and spoken phonemes \cite{Agarwal2020detecting} or cues related to the specific words uttered by the identity \cite{Agarwal2021watch} or inconsistencies between  inner and outer face regions \cite{dong2022protecting}. These methods present some limitations: in particular, they rely on video-only features, sometimes complemented by categorical information, neglecting precious audio information. Moreover, they often need several hours of videos of the identity under test.

In this work, we propose a new person-of-interest (POI) deepfake detector, called \NAME. The key feature of our approach is the use of a multi-modal analysis. More specifically we rely on a contrastive learning approach and train an audio and video network so that the learned representation characterizes temporal segments of the same identity close to one-another but far from each-other for different identities, as can be seen in Figure \ref{fig:teaser}.
At test time, we compute similarity indices between the features extracted from the video under analysis and those extracted by a set of POI-based reference videos. We also include joint audio-video similarity indices that have been shown to improve the discrimination ability of the detector.
Overall the proposed method ensures a number of important benefits:

\vspace{2mm}
\noindent
\bb{Generalization.}
Since training does not rely on fake videos, our detector works equally well on known and unknown manipulations. This ensures good generalization to attacks not seen during training (as none are seen during training). The detector can deal with any manipulation method (face swapping, facial reenactment or anything else) with performance that depends only on the fidelity of the video, not on specific inconsistencies or artifacts of the manipulation method.

\vspace{2mm}
\noindent
\bb{Flexibility.}
Due to our multi-modal approach, we can detect also video-only and speech-only manipulations, and even the swapping of a real audio track on the real video of the original identity. When the manipulation involves video and audio jointly, the joint-modality analysis improves performance.

\vspace{2mm}
\noindent
\bb{Robustness.}
The detector is robust to many challenging conditions frequently encountered in real scenarios, where videos are compressed or even maliciously attacked.

\vspace{2mm}
\noindent
\bb{No need of re-training.}
Training does not require videos of the POI under test, hence re-training is not needed when testing new identities, and only a few short POI videos (around 10 minutes) are necessary at test time.

\vspace{2mm}
\noindent
To summarize, our main contributions are the following:
\begin{itemize}
\item 
We propose \NAME, an audio-visual deepfake detection approach which learns a person-of-interest based representation and exploits single and multi-modality consistencies on video temporal segments.
\item 
Experiments show that our method beats SOTA approaches of a significant margin, especially in the most challenging scenarios of compressed and adversarially attacked videos with AUC and accuracy both improving in the range 7\%-14\%.
\end{itemize}

\section{Related Work}

\paragraph{Single-modality methods.}
In the large and rapidly growing literature on deepfake detection, most methods rely on supervised training, leveraging large datasets of real and fake videos. For the majority, they analyze only video and exploit low-level features, artifacts due to some imperfections of the generation method. These methods are typically very effective when the target video has been generated with a manipulation present in training, and almost useless otherwise \cite{Verdoliva2020media,Tolosana2020DeepFakes}. Since new methods for generating synthetic data are proposed by the day, this latter case may occur quite frequently. 
Even assuming that examples of new types of manipulation were available, re-training the models on datasets that grow without bounds becomes intractable. On the other hand, fine-tuning them on the new data would likely disrupt performance of old ones (catastrophic forgetting). To face this problem, specific solutions have been proposed in the literature, resorting to few-shot learning \cite{Cozzolino2018,Jeon2020TGD} or incremental learning \cite{Marra2019incremental,Khan2021video} approaches. In any case, the problem remains of readily acquiring examples of new manipulations.
Another simple, yet effective, resource to improve generalization is augmentation. 
For forensic applications, beyond the standard operations considered in computer vision, it is important to include compression and resizing, so as to gain robustness against the typical impairments caused by social networks. In addition, some dedicated forms of cut-out appear to be useful for deepfake detection \cite{Das2021towards}.
Ensembling is also helpful to boost the performance, and to gain robustness against possible misalignments \cite{Bonettini2020video}. 

Aside from limited performance, another significant limitation of the above approaches is the lack of interpretability. This problem is partially addressed by methods that aim at detecting some specific cues related to the generation process. One direction is to rely on low-level artifacts caused by the up-sampling operation, clearly visible in the Fourier domain. Accordingly, frequency-based analyses are carried out in \cite{Li2021frequency,Luo2021generalizing,Liu2021spatial}. Another direction is to learn the specific artifacts introduced by blending, which is a necessary processing step in many different manipulation approaches \cite{Li2020face}. More in general, attention mechanisms can be used to guide the network to focus on low-level and/or high-level inconsistencies, both in the spatial and in the temporal domain \cite{Zhao2021multiattentional,Dang2020detecting,Zhao2021learning,Wang2021representative,Zheng2021exploring,Zhu2021face}.
While these solutions provide some more insight on the type of manipulation performed, they are also very vulnerable to all quality impairment actions that disguise the low-level features they rely upon. These include not only casual image impairment processes, but also voluntary perturbations and adversarial attacks \cite{Hussain2021adversarial,Neekhara2021adversarial} which are becoming more and more commonplace.
To gain robustness, the method proposed in \cite{Haliassos2021lips} is based on semantic features by focusing on inconsistencies in the mouth movements. The spatio-temporal architecture is pre-trained on the visual speech recognition task and then fine-tuned on mouth embeddings of real and forged videos.

\paragraph{Multimodal analysis.}
In recent years, a few pioneering works have began analyzing audio and video jointly to perform deepfake detection. Some works look for inconsistencies between the audio and video content. The method developed in \cite{Korshunov2018speaker,Korshunov2019tampered}, for example, relies on the inability of some generation methods to correctly synchronize the audio stream with the video content. Likewise, the key idea of \cite{Zhou2021joint} is to learn and exploit the intrinsic synchronization between video and audio. However, as technology has advanced quickly, there are now several methods that can generate realistic deepfakes where speech and lips movements are accurately synchronized \cite{Prajwal2020lip}, making audio-visual synchronization analysis extremely challenging.
The approach proposed by \cite{Mittal2020} focuses on the extraction of the emotion features from both modalities together with a similarity analysis within the same audio and the same video.
In \cite{Wang2021M2TR}, a multi-modal and multi-scale transformer is designed to exploit both spatial and frequency domain artifacts, while in \cite{Chugh2020notmade} the idea is to look for inconsistencies between audio and visual streams by training a modality dissonance score.
While these methods show promising results, they require both fake and real videos for training, which could impair their generalization ability.

\begin{figure*}[t]
    \centering
    \includegraphics[page=1, width=0.8\linewidth, trim=0 90 0 0]{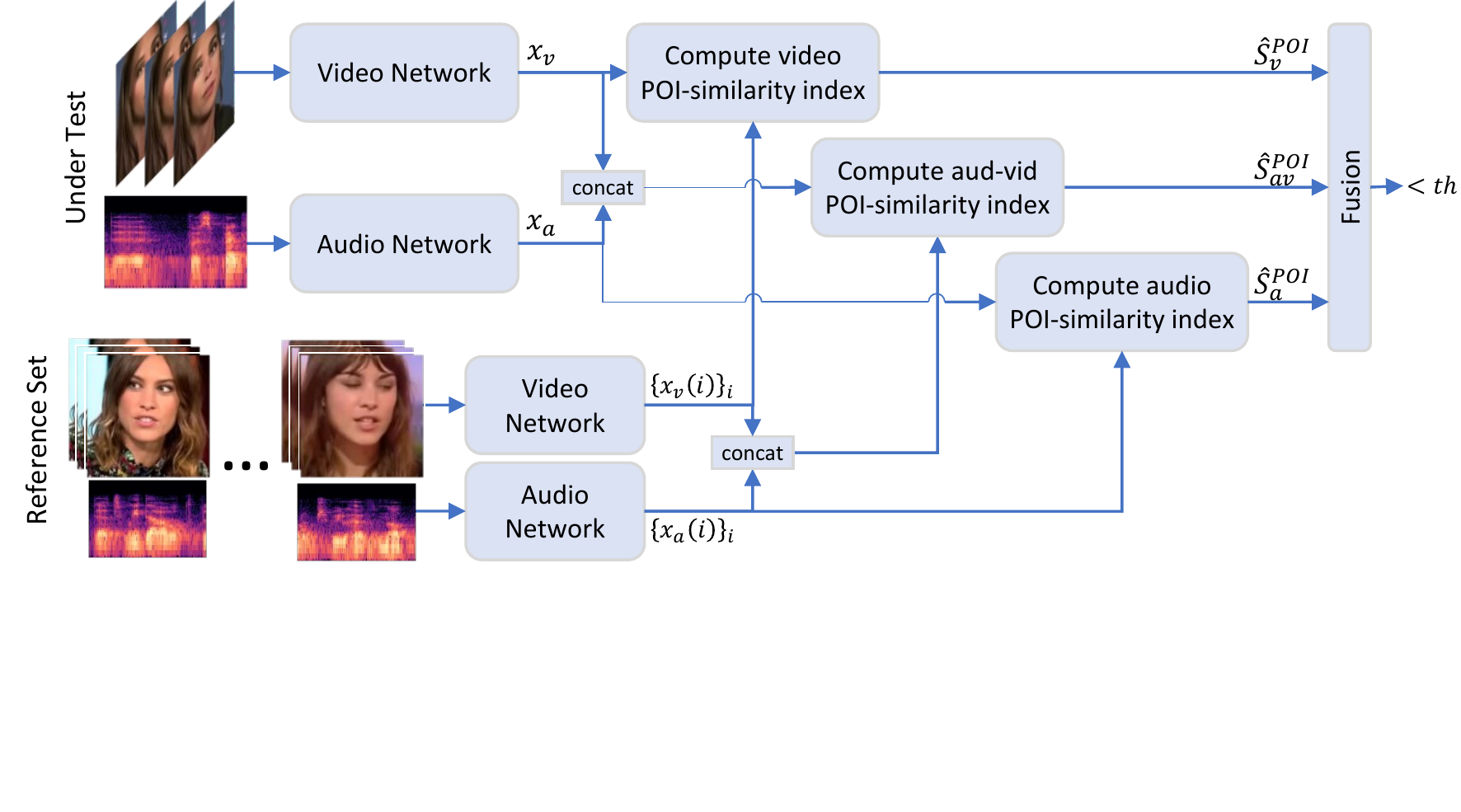}
    \caption{\NAME~testing scheme. We extract from the audio and video segments the embedded vectors and compare them with those extracted from a set of pristine videos of person of interest by means of the POI similarity indices (audio-only, video-only, audio-video). Finally, a fusion POI similarity index is computed.}
    \label{fig:testing_scheme}
\end{figure*}

\paragraph{POI-based detection.}
To avoid being polarized by the type of manipulated videos included in the training set, one may choose to learn only an intrinsic model of pristine videos, and interpret all deviations from this model as an anomaly and, therefore, a manipulation \cite{Cozzolino2019extracting, haliassos2022leveraging}. In particular, the biometrics of an individual allow for the construction of expressive and reliable models \cite{Agarwal2019}. Under this perspective, deepfake video detection can be interpreted as a POI-based attribution task. In fact, the question to answer is thus not, {\em real or fake?}, but rather {\em is this the POI or not?}. This identity verification approach is very general as it can handle both cheapfakes and deepfakes at the same time. 

Several works in this area rely on facial and head movement features \cite{Agarwal2019,Cozzolino2021idreveal,bohavcek2022protecting}. Specifically, in \cite{Agarwal2020} a method is proposed that includes a face recognition module to capture facial features
and another module to learn frame-based facial movements and expressions, that can better measure spatio-temporal biometric behaviors. This approach turns out to be very effective on face-swapping manipulations, much less on facial reenactment ones. To handle both types of attacks, in \cite{Cozzolino2021idreveal} a low-dimensional 3D morphable model of the person is extracted from the target video so as to analyze the face motion by means of an adversarial learning strategy. In \cite{Agarwal2020detecting}, inconsistencies between the mouth shape dynamics and the spoken phonemes are exploited, and the idea is further developed in \cite{Agarwal2021watch} where inconsistencies between facial movements and spoken words are targeted. This approach can handle also sophisticated speech manipulations where facial characteristics have not been altered. Its main drawback is the need to train on several hours of videos of the POI, which may not be always available and so reduces flexibility. More recently, in \cite{dong2022protecting} it is proposed an identity-based method that detects if the inner and outer face regions belong to the same individual. 

\section{Method}
\label{sec:method}

We propose a new method for deepfake detection which exploits audio-visual features of the portrayed individual.
Our key assumption is that any manipulation method, however accurate, will eventually perturb some of these features, which will be thus identified as anomalous, allowing for the detection of the attack.
So, our method does not look for direct traces of manipulation, such as generation artifacts, but rather for the indirect but compelling clue that the individual present in the video is not the person who is claimed to be.
These features are learned in a suitable embedding space by means of a contrastive learning procedure, looking for consistencies/inconsistencies among embedded vectors associated with different identities.
Our method trains exclusively on real videos, a large dataset comprising more than 5,000 identities with the associated audio \cite{Chung2018voxceleb2}, thereby maintaining independence from any specific manipulation and an intrinsic high generalization ability.
The dataset presents around 61\% males and spans a wide range of different ethnicities and ages.
It varies in terms of environment (indoor/outdoor), hence including different lighting
and variations in pose.

Fig.~\ref{fig:testing_scheme} shows the testing scheme.
From each 3-second segment of the video at test time, two 256-component embedded vectors are extracted by means of dedicated neural networks (in both cases, ResNet-50 with Group-Normalization \cite{Wu2018group}), one for the video modality and one for the audio modality.
The input of the audio network is the $300\times257$ amplitude spectrogram extracted from the audio using a window length of $25$ ms and a stride of $10$ ms.
The input of video network is the cropped faces from $25$ frames evenly distributed along the segment.
The output vectors are denoted as $x_{m}(c)$, with $c$ indicating the video-segment and $m \in \left\{a,v\right\}$ the modality ($a$ for audio and $v$ for video).

During training, performed by the contrastive learning approach described in Section \ref{sec:learning}, 
the audio and video networks learn to extract features that are generally close to one-another for segments of the same identity and far from each-other for segments of different identities.
Accordingly, in the testing procedure, described in Section \ref{sec:testing}, we compute POI similarity
indices between the features extracted from the target video and those extracted by a set of reference videos of the same POI.
In addition, joint audio-video similarity indices are computed as they are found experimentally to improve performance.
In the following, we will describe in detail both train and test phases.

\begin{figure*}
    \centering
    \includegraphics[page=9, width=0.8\linewidth, trim=0 114 0 0]{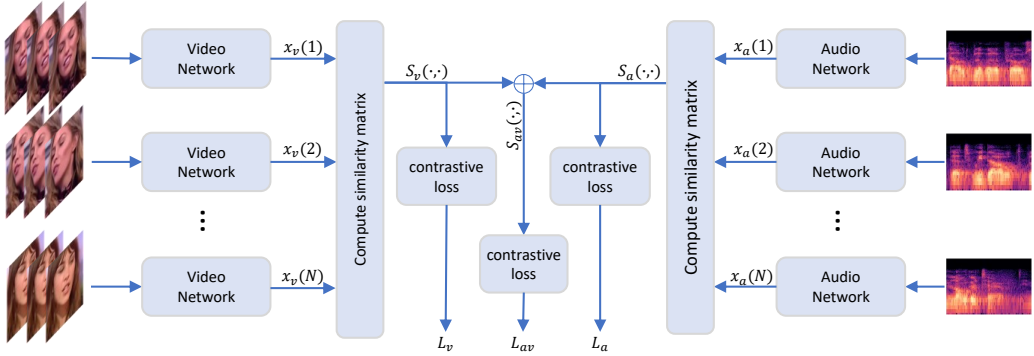}
    \caption{\NAME~training scheme. At each training iteration, we analyze $N$ video-segments and extract the embedded vectors from the audio and video signals. All the embedded vectors are compared computing three $N \times N$ matrices of similarity measures for only-video, only-audio and audio-video, respectively. For each matrix, a contrastive loss is evaluated to push closer the embedded vectors of the same individual but move farther from those of different individuals.}
    \label{fig:schema_training}
\end{figure*}

\subsection{Contrastive Learning}
\label{sec:learning}

Our model is trained using a contrastive learning formulation to embedded vectors of a video to be close to embedded vectors of the same identity, but far from those of different identities.
Overall the loss is defined as:
\begin{equation}
    \mathcal{L}_{tot} =  \mathcal{L}_{v} + \mathcal{L}_{a} + \lambda \, \mathcal{L}_{av},
    \label{equ:main_loss}
\end{equation}
where $\mathcal{L}_{v}$ and $\mathcal{L}_{a}$ are the contrastive losses applied to the single modality, while $\mathcal{L}_{av}$ is a joint contrastive loss that takes into account both modalities. Note that the joint contrastive loss is used to push a modality to make up for the shortcomings of the other modality, this gives an improvement when both the modalities are used in testing.

For all three terms, we adopt a multi-way matching loss that compares the positive matches with all the negative matches, leading to a more stable learning as compared with popular pairwise losses, such as the triplet loss, 
that compare each positive match with only one negative match \cite{Sohn2016improved}.
We adopt the contrastive loss function used in \cite{Khosla2020supervised,Cozzolino2021idreveal}, where the similarity between feature vectors is evaluated as the negative of squared Euclidean distance normalized by a factor $\tau$.
The contrastive loss function is expressed as:
\begin{equation}
    \mathcal{L}_{m} =  - \sum_{c} \log
    \frac{\sum_{k \in \mathcal{N}_c}  e^{S_{m}(c,k)}}
         {\sum_{k\neq c} e^{S_{m}(c,k)} } \hspace{6mm} m \in \left\{a,v,av\right\},
    \label{equ:contrastive}
\end{equation}
\noindent
where $c$ and $k$ are 3-second segments extracted from different videos,
and $S_{m}(c,k)$ measures their similarity according to modality $m \in \left\{a,v,av\right\}$.
Note that the summation at the numerator is restricted to the set $\mathcal{N}_c$,
comprising only segments $k$ that portray the same identity as $c$.
The single-modality similarity measures are defined as:
\begin{equation}
    S_{m}(c,k) = -\frac{1}{\tau} \left\| x_{m}(c) - x_{m}(k) \right\|^2 \,\,\,\,\,\,\,\,\,\,\, m \in \left\{a,v\right\}
\end{equation}
while the joint similarity measure $S_{av}(c,k)$ is the sum of the two single-modality measures. 

We divide the dataset on a per-individual basis, using $4608$ identities for training and $512$ for validation.  
Note that we excluded the 500 identities that were used to build the deepfake dataset FakeAVCelebV2 \cite{Khalid2021fakeavceleb} in order to avoid any possible polarization in our analysis.
The networks are trained for $12$ epochs with $2304$ batches per epoch, and each batch is formed by $8\times 8$ video-segments, with $8$ segments each for $8$ different individuals.
We use the Adam optimizer with decoupled weight decay \cite{Loshchilov2017decoupled}, a learning rate of $10^{-4}$, weight decay of $0.01$, and the $\tau=0.01$.
To increase robustness we use a wide variety of augmentations, including geometric operations (rescaling, rotation, flipping and shifting), point-wise operations (random variation of brightness, contrast, saturation, and hue) and image impairments (blurring, compression, noise adding and patch removal).

\subsection{Testing}
\label{sec:testing}

We assume to have a reference set of at least $10$ pristine videos of the person of interest, 
each one lasting about $30$ seconds, for a total of $100$ disjoint segments, that is a reference set $\mathcal{R}$.
At test time,
for each segment, $c$, of the target video and each modality
we compute an index,
which measures the similarity between the test segment and the
closest reference segment:
\begin{equation}
    S^{\rm POI}_m(c) = \underset{i\in \mathcal{R}}{\max} \, S_{m}(c,i) \hspace{6mm} m \in \left\{a,v,av\right\},
\end{equation}
These statistics will guide our decision, as they provide an indication of how plausible it is 
that the segment portrays the intended POI.
Indeed we rely on the normalized metrics defines as
\begin{equation}
    \widehat{S}^{\rm POI}_m(c) = \frac{S^{\rm POI}_m(c)-\mu^{\rm POI}_m}{\sigma^{\rm POI}_m}
    \label{equ:poi_index}
\end{equation}
where $\mu^{\rm POI}_m$ and $\sigma^{\rm POI}_m$ are, for each modality and person, the mean and standard deviation of the similarity index estimated from the pristine videos of the reference set.
In detail, these are estimated on the set $\left\{ S^{\rm POI}_m(i) \right\}_{i \in \mathcal{R}}$ computed by avoiding to evaluate the similarity measures between two segments of the same video.
For our assumptions on the reference set, 
there is a pretty large number of similarity indices to average upon, so this estimates can be considered reliable.
We also consider a fusion POI-similarity index, motivated by preliminary experiments, as the minimum of the three normalized POI-similarity indices.
Then, under the hypothesis that the video under test is real, that is, it portrays the claimed POI,
the normalized similarity index can be regarded as a standard Gaussian random variable
 $   \widehat{S}^{\rm POI}_m(c) \sim {\cal N}(0,1)$.
This modeling allows one to compute the probability of false alarm, $P_{fa}$, in closed form for any decision threshold or, conversely, to set the decision threshold so as to obtain the desired false alarm rate.
In the experimental part, we will fix the threshold considering a desired false alarm rate of 10\%. 
Considering a whole target video of, say, 30 seconds, we have 10 disjoint segments with 10 POI-similarity indices for each modality, certainly not independent of one another, to be taken into account jointly.
Therefore, we calculate the overall decision statistic by averaging the POI-similarity indices of all the disjoint segments in the target video.

\begin{table}[t!]
\small
\centering
\setlength{\tabcolsep}{1pt}
\scalebox{0.9}{
\begin{tabular}{lK{0.2cm}K{0.2cm}K{0.2cm}K{0.02cm}K{0.9cm}K{0.9cm}K{0.9cm}K{0.9cm}K{0.02cm}K{0.9cm}K{0.9cm}K{0.9cm}K{0.9cm}}
\toprule
&            &            &             && \multicolumn{4}{c}{$\lambda=0$} & & \multicolumn{4}{c}{$\lambda=1$} \\
&            &            &             && video & audio &    AV & Fusion  & & video & audio &    AV & Fusion  \\
&        $v$ &        $a$ &        $ai$ && index & index & index &         & & index & index & index &         \\ \cmidrule{1-4} \cmidrule{6-9} \cmidrule{11-14}

\multirow{2}{*}{\rotatebox{90}{AUC (\%)~~~}}
& \checkmark &            &             &&  79.0 &  49.5 &  70.6 &    73.4 & &  78.3 &  49.1 &  68.8 &    71.4 \\
& \checkmark &            & \checkmark  &&  74.4 &  97.0 &  93.5 &    95.5 & &  73.1 &  96.0 &  94.6 &    96.0 \\
&            & \checkmark & \checkmark  &&  53.6 &  93.3 &  82.0 &    88.7 & &  49.5 &  95.2 &  83.2 &    90.9 \\
& \checkmark & \checkmark & \checkmark  &&  72.0 &  99.0 &  95.4 &    96.5 & &  70.8 &  98.8 &  96.4 &    97.5 \\ \cmidrule{2-4} \cmidrule{6-9} \cmidrule{11-14}
& \multicolumn{3}{c}{AVG}               &&  69.8 &  84.7 &  85.4 &    88.5 & &  67.9 &  84.8 &  85.8 & \textbf{88.9} \\\cmidrule{1-4} \cmidrule{6-9} \cmidrule{11-14}

\multirow{2}{*}{\rotatebox{90}{Pd@10\% (\%)~~}}
& \checkmark &            &             &&  50.1 &  15.1 &  29.9 &    39.4 & &  40.9 &  14.6 &  34.1 &    37.3 \\
& \checkmark &            & \checkmark  &&  49.6 &  94.1 &  87.9 &    90.9 & &  46.5 &  94.7 &  91.6 &    94.5 \\
&            & \checkmark & \checkmark  &&  15.6 &  89.1 &  62.5 &    65.6 & &  10.9 &  90.6 &  64.1 &    76.6 \\
& \checkmark & \checkmark & \checkmark  &&  42.3 &  98.8 &  90.0 &    95.6 & &  37.0 &  97.6 &  94.3 &    95.6 \\\cmidrule{2-4} \cmidrule{6-9} \cmidrule{11-14}
& \multicolumn{3}{c}{AVG}               &&  39.4 &  74.3 &  67.6 &    72.9 & &  33.8 &  74.4 &  71.0 & \textbf{76.0} \\
\bottomrule
\end{tabular}
}
\vspace{1mm}
\caption{Results in terms of AUC and Pd\%. We compare the four similarity indices and evaluate the effect to include the joint contrastive loss term during training.}
\vspace{-5mm}
\label{tab:abl}
\end{table}

\section{Experimental Results}
\label{sec:results}

\subsection{Datasets and Metrics}

To evaluate our method and reference state-of-the-art approaches,
we consider several deepfake datasets that also provide identity information:

\vspace{2mm} \noindent
\bb{pDFDC}, preview DeepFake Detection Challenge dataset \cite{Dolhansky2019preview}.
It includes realistic face-swapping manipulations relative to $68$ identities.

\vspace{2mm} \noindent
\bb{DF-TIMIT}, DeepFake-TIMIT \cite{Korshunov2018deepfakes}.
It contains face-swapping manipulations applied to $320$ videos of $32$ identities of the Vid-TMIT dataset \cite{Sanderson2009multi}.

\vspace{2mm} \noindent
\bb{FakeAVCelebV2}, Audio-Video Deepfake dataset ~\cite{Khalid2021fakeavceleb}.
It comprises both face-swapping and facial reenactment methods.
To generate cloned fake audio a transfer learning-based real-time voice cloning tool (SV2TTS \cite{Jia2018transfer}) is used.

\vspace{2mm} \noindent
\bb{KoDF}, a large-scale Korean DeepFake dataset \cite{Kwon2021kodf}.
It contains $403$ identities and is composed of manipulated videos generated by six different techniques, including both face swapping and face-reenactment manipulations.

\vspace{2mm} \noindent
We report results using the following metrics: AUC, which is very popular in the literature and does not require to set a threshold, Accuracy, with a fixed 0.5 threshold  and the probability of detection obtained for a 10\% false alarm rate (Pd@10\%).
All these metrics are evaluated at video-level for each technique included in the analysis.

\subsection{Ablation Study}

\noindent
\bb{Influence of similarity indexes.}
First, we assess the importance of each of the similarity indices, $\{a,v,av\}$, contributing the decision statistic and, 
also, the importance of including or not the joint contrastive loss term: $\lambda = 1$ or $0$ in eq.~\ref{equ:main_loss}. 
To this end, we experiment on a subset of FakeAVCelebV2 and KoDF datasets comprising a total of more than 140 individuals, and the whole variety of manipulations.
Results are presented in Tab. \ref{tab:abl} in terms of AUC (top) and Pd@10\% (bottom).
For this analysis, the deepfakes are grouped according to which parts of the multimedia asset (video+audio) is manipulated:
only the video, only the audio, both video and audio.
Actually, the first group is further divided in two subgroups depending on whether the non-manipulated audio is 
consistent with the claimed identity or not (for example, because only the face was swapped).
In the first case, the audio similarity index will not help, in the second case it could be important.
These four groups are identified by the checkmarks in the first three columns of the table,
indicating video manipulation ($v$), audio manipulation ($a$), audio inconsistency ($ai$).

Let us focus, for the time being, on the $\lambda=0$ case and the AUC measure (upper-left part of the table).
Several important observations are already possible.
First, using the audio similarity index ensures a large performance gain with respect to using video-only features, from 15\% to 20\% on the average.
This is a huge improvement, and hence a very significant result.
We believe this is likely due to the limited emphasis that the audio component has received to date with respect to the video component.
As expected, the audio index becomes useless in case 1, but the video index makes up for this loss.
Notably, in this case, including the audio index in the decision statistic is not immaterial, 
as it reduces the performance in both the audio-video and fusion cases.
On the average, however, the fusion of all three indices grants a significant gain with respect to all other cases.
The best performance is observed in case 4, of course, when all components are manipulated. 
Moving right to the $\lambda=1$ case, where the joint contrastive loss term is activated, we observe a limited but consistent performance improvement, all other considerations remaining the same.
Moving down, all numbers drop significantly, as the Pd@10\% is a much less forgiving performance measure, but general behaviors are basically confirmed.
Notable differences concern the poor performance obtained when only the video index is used, 
and the great improvement achieved by including the contrastive loss term, especially with the fusion strategy.

\begin{figure}[t!]
    \centering
    \includegraphics[page=5, width=1.0\linewidth, trim=20 122 20 0, clip]{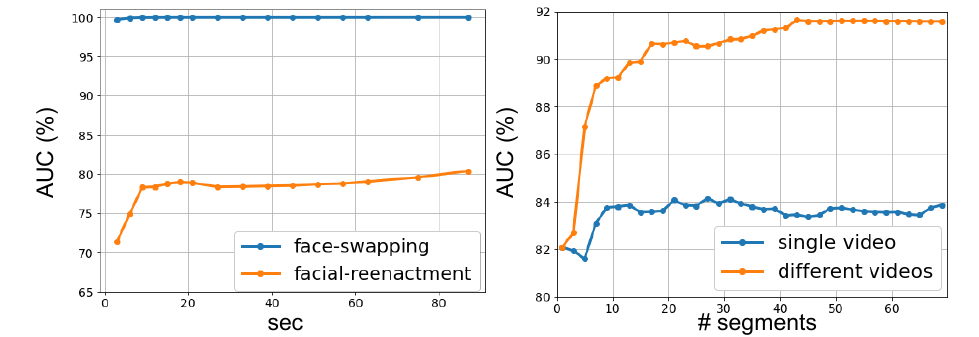}
    \caption{Results in terms of AUC by varying the length of the video under test (left) and by varying the number of video segments of the reference set (right).}
    \label{fig:videolength}
\end{figure}

\begin{figure}[t!]
    \centering
    \includegraphics[page=3, width=1.\linewidth, trim=0 60 0 0]{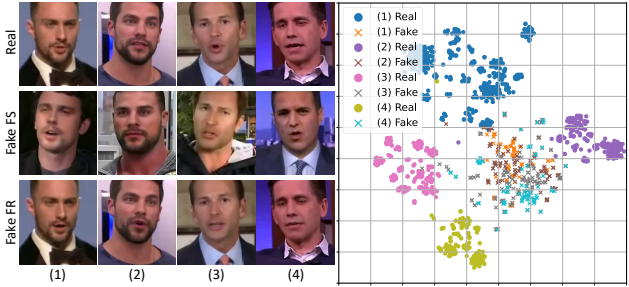}
    \caption{The t-SNE visualizations of features extracted from real and fake videos of four individuals. Each dot represents the feature relative to a video segment.}
    \label{fig:tSNE}
\end{figure} 

\vspace{2mm}
\noindent
\bb{Influence of test video length.}
We aim at studying the role of the test video length, 
analyzing how the performance varies as a function of the number of available segments.
Results are shown in the left graph of Fig.\ref{fig:videolength}, separately for face-swapping (FS) and facial reenactment (FR) manipulations. 
Interestingly, for face-swapping, even a single 3-second snippet is sufficient to reach perfect detection. 
Detecting facial reenactment is more challenging, as well known, with an AUC that starts at 70\% with a single segment
and grows slowly with the size of the video to reach an 80\% plateau with about 20 seconds of video.
  
\vspace{2mm}
\noindent
\bb{Influence of size and variety of the reference set.}
Results are shown in the right graph of Fig.\ref{fig:videolength} and are  averaged on all types of manipulations.
It appears that a single POI video is not sufficient to achieve good performance, no matter how long the video is, with an AUC always at 84\% or below.
Instead, using multiple videos, the AUC improves rapidly beyond 90\% to reach 92\% when at least 40 segments are available.
In summary, variety seems very important, more than sheer data size.
In Fig.\ref{fig:videolength}, we observe that results on FR manipulations are worse than on FS ones, since the former better preserves the characteristics of the identity, and so they are more challenging for our approach.

\vspace{2mm}
\noindent
\bb{t-SNE representation.}
In Fig.\ref{fig:tSNE}, we show a scatter plot of the 2D projections, obtained by means of the t-SNE \cite{Van2008visualizing} dimensionality reduction technique,
of some selected embedded vectors (concatenation of audio and video features). 
The vectors are extracted from real (circles) and fake (crosses) videos of the four identities shown on the left part of the figure.
These were selected so as to present rather similar features in order not to induce systematic biases in the projections.
Even so, embedded vectors relative to different individuals appear to from relatively compact clusters, well separated from one another.
Moreover, with few exceptions, their manipulated counterparts turn out to be quite distant from the corresponding real videos.

\subsection{Comparative Performance Analysis}

\vspace{2mm}
\noindent
\bb{State-of-the-art approaches.}
We consider approaches for which publicly available implementations are available.
The methods used for our comparison are ensemble frame-based methods methods: Seferbekov~\cite{Selimsef2020}, Real Forensics~\cite{haliassos2022leveraging}, temporal-based methods: FTCN (Fully Temporal Convolution Network)~\cite{Zheng2021exploring}, LipForensics~\cite{Haliassos2021lips}, audio-visual methods: MDS-based~FD~\cite{Chugh2020notmade}, Joint AV \cite{Zhou2021joint}, and identity-based ones: ICT (Identity Consistency Transformer)~\cite{dong2022protecting}, ID-Reveal~\cite{Cozzolino2021idreveal}.
A detailed description of these approaches can be found in the
supplemental document. 

\begin{table*}[t]
  {\small
    \centering
    \scalebox{0.94}{
     \begin{tabular}{llCCCCC}
    \toprule 
\ru & AUC/ACC    & pDFDC     & DF-TIMIT        &  FakeAVCel. &        KoDF & AVG \\ \toprule
\multirow{10}{*}{\rotatebox{90}{High quality}}
\ru & Seferbekov       & ~85.5 /~72.0 & ~95.7 /~87.6 & ~98.6 /~95.0 & ~88.4 /~79.2 & ~92.0 /~83.5 \\ 
\ru & FTCN             & ~72.3 /~63.9 & 100.0 /~87.4 & ~84.0 /~64.9 & ~76.5 /~63.0 & ~83.2 /~69.8 \\ 
\ru & LipForensics     & ~68.7 /~60.0 & ~98.8 /~78.0 & ~97.6 /~83.3 & ~92.9 /~56.1 & ~89.5 /~69.3 \\ 
\ru & Real Forensics   & ~85.2 /~70.4 & 100.0 /~99.5 & ~88.3 /~76.2 & ~95.6 /~63.1 & ~92.3 /~77.3 \\
\ru & MDS-based~FD     & ~98.0 /~93.4 & ~56.6 /~55.8 & ~64.7 /~61.7 & ~65.3 /~63.1 & ~71.3 /~68.5 \\ 
\ru & Joint AV         & ~53.2 /~52.8 & ~50.0 /~50.0 & ~55.1 /~48.6 & ~49.4 /~49.2 & ~51.9 /~50.1 \\
\ru & ICT              & ~77.1 /~70.7 & ~87.8 /~77.1 & ~68.2 /~63.9 & ~62.5 /~58.9 & ~73.9 /~67.7 \\ 
\ru & ICT-Ref          & ~87.6 /~79.8 & 100.0 /~94.7 & ~71.9 /~64.5 & ~79.4 /~60.3 & ~84.7 /~74.8 \\ 
\ru & ID-Reveal        & ~91.3 /~80.4 & ~99.0 /~92.8 & ~70.2 /~60.3 & ~87.6 /~63.7 & ~87.0 /~74.3 \\
\ru & \NAME~(ours)     & ~95.2 /~86.7 & ~99.2 /~85.7 & ~94.1 /~86.6 & ~89.9 /~81.1 & \textbf{94.6}/\textbf{ 85.0} \\ 
\toprule
\multirow{10}{*}{\rotatebox{90}{Low quality}}
\ru & Seferbekov       & ~62.6 /~54.0 & ~81.8 /~71.4 & ~61.7 /~58.5 & ~79.6 /~55.9 & ~71.4 /~59.9 \\ 
\ru & FTCN             & ~51.3 /~50.0 & ~78.3 /~58.9 & ~37.6 /~42.1 & ~68.4 /~58.6 & ~58.9 /~52.4 \\ 
\ru & LipForensics     & ~46.5 /~45.8 & ~70.9 /~62.1 & ~58.3 /~53.8 & ~85.7 /~52.3 & ~65.3 /~53.5 \\ 
\ru & Real Forensics   & ~55.8 /~57.1 & ~81.1 /~73.6 & ~52.9 /~49.2 & ~88.0 /~56.3 & ~69.4 /~59.0 \\ 
\ru & MDS-based~FD     & ~95.6 /~90.0 & ~52.6 /~52.2 & ~61.1 /~58.6 & ~64.6 /~62.4 & ~68.7 /~65.8 \\ 
\ru & Joint AV         & ~53.4 /~52.5 & ~51.2 /~50.2 & ~55.2 /~48.4 & ~50.4 /~49.6 & ~52.6 /~50.2 \\
\ru & ICT              & ~72.0 /~66.4 & ~84.4 /~74.1 & ~66.9 /~61.5 & ~61.1 /~59.3 & ~71.1 /~65.3 \\ 
\ru & ICT-Ref          & ~82.4 /~73.8 & 100.0 /~96.5 & ~71.2 /~59.6 & ~78.1 /~62.3 & ~82.9 /~73.1 \\ 
\ru & ID-Reveal        & ~86.5 /~73.9 & ~93.6 /~75.5 & ~70.8 /~58.9 & ~85.0 /~63.8 & ~84.0 /~68.0 \\ 
\ru & \NAME~(ours)     & ~93.9 /~82.0 & ~98.2 /~76.3 & ~94.4 /~88.7 & ~89.0 /~81.5 & \textbf{~93.9}/\textbf{~82.1} \\ 
    \bottomrule
    \end{tabular}
    }
    \vspace{1mm}
    \caption{Comparison with state-of-the-art on pDFDC, DeepfakeTIMIT, FakeAVCelebV2, and KoDF. 
    Results are shown on high-quality (HQ) videos and low-quality (LQ) ones.}
    \label{tab:comparison}
    }
\end{table*}

\begin{table*}[t]
\setlength{\tabcolsep}{1pt}
  {\small
    \centering
    \scalebox{0.95}{
     \begin{tabular}{lK{1.2cm}K{1.2cm}K{1.2cm}K{1.2cm}K{1.2cm}K{1.2cm}K{1.2cm}K{1.2cm}K{1.2cm}K{1.2cm}}
    \toprule 
\ru         & Sef. & FTCN & Lip For. & Real For. & MDS-FD & Joint AV & ICT & ICT-ref & ID-Ref. & POI-For. \\ \midrule
\ru AUC     & 61.5 & 58.3 & 54.8 & 55.5 & 55.4 & 47.1 & 61.5 & 78.0 & 73.4 & \textbf{80.5}\\
\ru Pd@10\% & 23.7 & 10.3 & 16.2 &  3.7 & 13.1 & 10.7 & 19.1 & 48.2 & 27.6 & \textbf{49.6} \\
\ru ACC     & 56.7 & 52.6 & 48.0 & 47.1 & 53.4 & 49.8 & 53.7 & 65.3 & 67.1 & \textbf{73.0} \\
    \bottomrule
    \end{tabular}
    }
    \vspace{1mm}
    \caption{Results on the attacked KoDF subset.}
    \label{tab:attack}
    \vspace{-5mm}
    }
\end{table*}
\vspace{2mm}
\noindent

\bb{Training.}
In order to ensure a fair comparison, Seferbekov, FTCN, LipForensics and Real Forensics are all trained on the FaceForensics++ dataset \cite{Roessler2019ff++}, which includes different types of facial manipulation (both FS and FR).
MDS-based~FD and Joint AV are trained on DFDC \cite{Dolhansky2020dfdc}, which includes also audio manipulations. ICT is trained on a publicly available data-set of pristine faces of 1 million identities, MS-Celeb-1M \cite{guo2016celeb}.
ID-Reveal, instead, is trained on VoxCeleb2 \cite{Chung2018voxceleb2} like our own method. In all cases, we ensure that training and test data do not overlap.
Note that the reference set for ICT-Ref, ID-Reveal and the proposal are built from the same set of pristine videos.

\vspace{2mm}
\noindent
\bb{Generalization analysis.}
Results are reported in Table \ref{tab:comparison} in terms of AUC and accuracy.
The top part of the table refers to high-quality (HQ) data, 
with videos compressed using H.264 coding with factor 23 and original audio tracks.
The bottom part is for low-quality (LQ) data, 
with videos compressed using H.264 coding with factor 40 and 25 fps, and audio tracks compressed using Advanced Audio Coding (AAC) with a sample-rate of 16K and a bit-rate of 48K.
As a preliminary observation,
note that all methods have much better figures in terms of AUC than Accuracy, 
confirming the difficulty in setting a global threshold and the need to properly calibrate each technique to make it work in realistic conditions.

On HQ videos, the proposed method outperforms by 2.3\% (AUC) and 1.5\% (Accuracy) on the average the best reference (Real Forensics and Seferbekov respectively) and more markedly the other methods.
The supervised audio-visual approach obtains very good performance on pDFDC since it was trained on the DFDC dataset, but poorly generalize to other datasets. 
Most competitive methods are Seferbekov and Real Forensics.
On LQ videos, however, the lead of POI-Forensics over all reference methods becomes very large, with an average gain of 11\% (AUC) and 12\% (Accuracy) over the second best.
Indeed, in this challenging scenario, only identity-based approaches keep providing a good performance.

\vspace{2mm}
\noindent
\bb{Robustness to adversarial attacks}.
Adding adversarial noise to an image can easily fool a classifier, and this has been shown to apply also to deepfake detectors \cite{Hussain2021adversarial,Neekhara2021adversarial}. 
In \cite{Kwon2021kodf} the fast gradient sign method \cite{Goodfellow2014explaining} has been used to simulate malicious attacks.
An EfficientNet-B4 network \cite{Tan2019efficientnet} with two fully connected layers has been trained as a deepfake detector and then attacked by means of adversarial examples. 
Our method and the selected reference ones were all tested on this dataset, obtaining the results reported in Table \ref{tab:attack}. 
As expected, all performance metrics reduce dramatically for all methods.
Only our method, and to a lesser extent ICT-ref and ID-Reveal, keep providing a reasonable performance in this scenario.
Note that for the sake of fairness, the proposed method does not use the audio information in this experiment, as it is not subject to the attack.

\section{Conclusions}

We introduced \NAME, an audio-visual deepfake detection method 
that is trained on real videos to learn a POI representation
useful to reveal if the identity of the video under test is the person that is claimed to be.
Our experiments show that by leveraging audio-visual information,
we can handle a wide variety of manipulations.
Most important, since our detector learns a real-world data model, it does not depend on patterns introduced by specific deepfake generators.
In addition, the multimodal analysis could be further enriched by including more information from other domains such as textual data.

\paragraph{Acknowledgment.}
We gratefully acknowledge the support of this research by a TUM-IAS Hans Fischer Senior Fellowship, a TUM-IAS Rudolf M\"o{\ss}bauer Fellowship and a Google Faculty Research Award.
This material is also based on research sponsored by the Defense Advanced Research Projects Agency (DARPA) and the Air Force Research Laboratory (AFRL) under agreement number FA8750-20-2-1004. 
The U.S. Government is authorized to reproduce and distribute reprints for Governmental purposes notwithstanding any copyright notation thereon. 
The views and conclusions contained herein are those of the authors and should not be interpreted as necessarily representing the official policies or endorsements, either expressed or implied, of DARPA and AFRL or the U.S. Government. 
In addition, this work has received funding by the European Union under the Horizon Europe vera.ai project, Grant Agreement number 101070093, and the ERC Starting Grant Scan2CAD (804724).
It is also supported by the PREMIER project, funded by the Italian Ministry of Education, University, and Research within the PRIN 2017 program.


\begin{appendix}
\section*{Supplemental Material}
In this appendix, 
we present additional ablation studies (Sec.\ref{ablation}) and more
results to prove the robustness capability of our method (Sec.\ref{robustness}).
Moreover, we briefly describe the state of the art methods we compare to, (Sec.\ref{methods})
and give some more details about the dataset used in the experiments (Sec.\ref{datasets}).
Finally, we summarize the limitations of our method (Sec.\ref{limitations}).

\section{Additional ablation studies}
\label{ablation}

In this Section, we conduct additional experiments to show that our approach
outperforms some state-of-the-art audio-visual speaker verification methods for both the person identification task and the deepfake detection task.
We consider three reference methods, all relying on a contrastive learning paradigm:
\begin{itemize}
\item
In SyncNet\footnote{available at \url{https://github.com/joonson/syncnet\_trainer}} \cite{Nagrani2020disentangled},
a robust speaker identity representation is proposed, based on an end-to-end self-supervised approach.
Specific constraints are imposed
on the identity, that should change slowly over time,
on the content, that instead should change quickly,
and on both factors that are enforced to be represented independently of one-another through a disentangling constraint.
\item
The approach proposed in \cite{Shon2019noise} learns an audio-visual embedding for person verification by using two separate networks for audio and video that are jointly trained.
Then, fusion is performed at feature-level, by relying on an attention mechanism that learns the salient modality of input data.
\item
In \cite{Sari2021multi} an audio-visual fusion system is also proposed where, after concatenating the two features,
fusion is performed by means of a multilayer perceptron.
\end{itemize}

\begin{table}[t]
  {\small
    \centering
    \scalebox{0.90}{
    \setlength{\tabcolsep}{2pt} 
     \begin{tabular}{rrK{1.2cm}K{1.2cm}K{1.2cm}K{1.2cm}K{1.2cm}}
    \toprule 
      & & 10 & 50 & 100 & 150 & 200  \\
\cmidrule{1-2} \cmidrule(lr){3-7}
\multicolumn{2}{c}{Random Clas.}
                                    &  10.0 & ~2.0 &  ~1.0 &  ~0.7 & ~0.5 \\
\cmidrule{1-2} \cmidrule(lr){3-7}
\multirow{2}{*}{~~audio}
 & \cite{Nagrani2020disentangled}   &  69.0 &  43.2 &  31.7 &  25.8 & 22.4 \\
 & ours                             &  81.0 &  67.6 &  61.1 &  54.3 & 50.1 \\
\cmidrule{1-2} \cmidrule(lr){3-7}
\multirow{2}{*}{~~video}
 & \cite{Nagrani2020disentangled}   &  53.0 &  32.8 &  24.8 &  20.6 & 18.4 \\
 & ours                             &  75.0 &  62.4 &  57.8 &  53.5 & 50.0 \\ 
\cmidrule{1-2} \cmidrule(lr){3-7}
 \multirow{4}{*}{~~both}
 & \cite{Nagrani2020disentangled}   &  82.0 &  56.6 &  45.0 &  39.9 & 35.9 \\
 & \cite{Sari2021multi}             &  \textbf{92.0} &  75.8 &  69.6 &  65.2 & 63.0 \\
 & \cite{Shon2019noise}             &  90.0 &  79.0 &  72.8 &  69.8 & 66.4 \\
 & ours                             &  91.0 &  \textbf{82.0} &  \textbf{77.9} &  \textbf{76.4} & \textbf{73.3} \\  
    \bottomrule
    \end{tabular}
    }
    \vspace{1mm}
    \caption{Results in term of ACC (\%) on the person identification task considering a variable number of identities, from 10 to 200.}
    \label{tab:person}
    }
\end{table}

\begin{table}[t]
  {\small
    \centering
    \scalebox{0.90}{
    \setlength{\tabcolsep}{2pt}
\begin{tabular}{lK{0.4cm}K{0.4cm}K{0.4cm}K{0.2cm}K{1.5cm}K{1.5cm}K{1.5cm}K{1.5cm}}
\toprule
&        $v$ &        $a$ &        $ai$ &&  \cite{Nagrani2020disentangled} & \cite{Sari2021multi} & \cite{Shon2019noise} & Ours \\ \cmidrule{1-4} \cmidrule{6-9}

\multirow{2}{*}{\rotatebox{90}{AUC (\%)~~~}}
& \checkmark &            &             && 66.3 & 61.0 & 60.8 &    71.4 \\
& \checkmark &            & \checkmark  && 80.5 & 93.4 & 94.7 &    96.0 \\
&            & \checkmark & \checkmark  && 91.0 & 73.6 & 80.4 &    90.9 \\
& \checkmark & \checkmark & \checkmark  && 93.8 & 93.7 & 95.1 &  97.5 \\ \cmidrule{2-4} \cmidrule{6-9} 
& \multicolumn{3}{c}{AVG}               && 82.9 & 80.4 & 82.8 & \textbf{88.9} \\\cmidrule{1-4} \cmidrule{6-9} 

\multirow{2}{*}{\rotatebox{90}{Pd@10\% (\%)~~}}
& \checkmark &            &             && 19.0 & 26.3 & 23.3 &    37.3 \\
& \checkmark &            & \checkmark  && 47.9 & 86.8 & 88.2 &    94.5 \\
&            & \checkmark & \checkmark  && 57.8 & 53.1 & 51.7 &    76.6 \\
& \checkmark & \checkmark & \checkmark  && 77.0 & 90.4 & 90.4 &    95.6 \\\cmidrule{2-4} \cmidrule{6-9} 
& \multicolumn{3}{c}{AVG}               && 50.4 & 64.1 & 63.4 & \textbf{76.0} \\
\bottomrule
\end{tabular}
}
\vspace{1mm}
\caption{Results in terms of AUC and Pd\%. We compare our approach with other different strategies of audio-visual POI identification considering several scenarios, i.e. four situations identified by the checkmarks in the first three columns, indicating a video manipulation ($v$), an audio manipulation ($a$) and an audio inconsistency ($ai$). }
\label{tab:abl2}
}
\end{table}

First, we study the person identification problem
using the VoxCeleb2 dataset \cite{Chung2018voxceleb2} and considering from 10 to 200 different identities
(not included in training).
For training, we collect 100 video segments for each subject. At testing time, we use 1-Nearest Neighbor classification.
Performance is evaluated in terms of accuracy on 10 video segments for each subject.
Results are shown in Table \ref{tab:person}.
Since our method and \cite{Nagrani2020disentangled} are able to provide results even using a single modality we include also these results, in the upper part of the table.
Then, in the lower part we report fusion-based results for all methods, where both audio and video are exploited.
In all cases, our method outperforms all references, with the only exception of fusion when only 10 identities are involved (where it is second best).
The performance gain is especially significant in the most challenging cases where a large number of identities are considered.
As an example, with 200 identities, the proposed method has an accuracy of 73.3\%, 7 points better than the second best.

For deepfake detection, results are reported in Table \ref{tab:abl2} in a similar way as in the main paper (see Section 4.2, Table 1).
We identify four groups according to the checkmarks in the first three columns, indicating video manipulation ($v$), audio manipulation ($a$), audio inconsistency ($ai$). 
The dataset used for the analysis is again a subset of FakeAVCelebV2 and KoDF, comprising a total of more than 140 subjects. 
To ensure a fair comparison, reference methods have been trained on the same dataset used for our approach.
Moreover, for the methods proposed in \cite{Shon2019noise,Sari2021multi} we used our backbone.  
Also in this scenario
our approach ensures a clearly superior performance with respect to all references
both in terms of AUC, with an average improvement of 6 percent points with respect to the second best,
and in terms of ${\rm Pd}@10\%$, in which case the average improvement reaches 12 percent points. 

\begin{table}
  {\small
    \centering
    \scalebox{0.90}{
    \setlength{\tabcolsep}{2pt} 
     \begin{tabular}{lK{1.6cm}K{1.6cm}K{1.6cm}K{1.6cm}}
    \toprule 
         & No & video-only & audio-only & both \\ 
         & noise & noise & noise & noise \\ \midrule
\ru AUC     & 89.9 & 78.8 & 75.4 & 74.2 \\
\ru Pd@10\% & 74.3 & 50.3 & 55.5 & 52.7 \\
    \bottomrule
    \end{tabular}
    }
    \vspace{1mm}
    \caption{Results of our method on the noisy KoDF subset. We added noise on video and audio and evaluated the performance considering video-only, audio-only and both.}
    \label{tab:noise_single}
    }
\end{table}

\begin{table*}
  {\small
    \centering
    \scalebox{0.90}{
     \begin{tabular}{lK{1.1cm}K{1.1cm}K{1.1cm}K{1.1cm}K{1.1cm}K{1.1cm}K{1.1cm}K{1.1cm}K{1.1cm}K{1.1cm}K{1.1cm}}
    \toprule 
\ru         & Seferbekov & FTCN & LipFor. & Real.For. &  MDS-based~FD & Joint AV & ICT &  ICT-Ref & ID-Reveal & ours \,\, (video)  \\ \midrule
\ru AUC     & 68.4 & 50.2 & 54.5 & 62.5 & 69.1 & 50.4 & 59.9 & 72.6 & 62.8 & \textbf{79.1}  \\
\ru Pd@10\% & 31.3 & 10.1 & 15.2 & 16.2 & 22.2 & 12.3 & 23.9 & 42.8 & 16.2 & \textbf{49.9}  \\
    \bottomrule
    \end{tabular}
    }
    \vspace{1mm}
    \caption{Results on the noisy KoDF subset. We compare our approach with Seferbekov~\cite{Selimsef2020}, FTCN (Fully Temporal Convolution Network)~\cite{Zheng2021exploring}, LipForensics~\cite{Haliassos2021lips}, RealForensics~\cite{haliassos2022leveraging}, MDS-based~FD~\cite{Chugh2020notmade}, ICT, ICT-Ref~\cite{dong2022protecting}, and
ID-Reveal~\cite{Cozzolino2021idreveal}.}
    \label{tab:noise}
    }
\end{table*}

\section{Additional robustness analysis}
\label{robustness}

In this Section, we provide some more insights on the robustness of our approach. 
In the main paper, we analyzed how compression and adversarial attacks impair the performance of our detector (Section 4.3, Table 2, and Table 3).
Here, we focus on Gaussian noise addition which
is well-known to strongly impair deepfake detection performance when only the video has been modified \cite{Haliassos2021lips}. 
Moreover, noise is a serious issue also for audio-based speaker verification \cite{Heo2020clova}.

In our experiments, we consider again a subset of the KoDF dataset \cite{Kwon2021kodf} 
and add Gaussian noise to both audio and video signals, 
with random intensity amounting to an SNR going from 5 to 15 dB for the audio signal, and a PSNR going from 13 to 27 dB for the video signal. 
In Table \ref{tab:noise_single}, we show the impact of noise addition on our detector.
We consider separately the cases where only the video, only the audio or both modalities are attacked.
In any case, the detector uses both modalities.
As expected, for such intense levels of noise, the performance reduces sharply.
Nonetheless, even in this scenario our method keeps ensuring a reasonable performance, with an AUC of around 74\%  when both audio and video are attacked.

This consideration is further reinforced by comparing results with state-of-the-art detectors (Table \ref{tab:noise}).
To ensure a fair comparison, we run our method using only the video signal, as all references do.
Even so, a large performance improvement is observed with respect to all competitors,
with over 8\% AUC gain and 16\% ${\rm Pd}@10\%$ gain with respect to the second best, ICT-Ref.

\section{State-of-the-art methods}
\label{methods}

In the main paper, we compare our proposal with six state-of-the-art approaches:

\vspace{2mm}
\noindent
\bb{Seferbekov~} \cite{Selimsef2020} is the first-place solution of the Kaggle Deepfake Detection Challenge \cite{Dolhansky2020dfdc}. It is based on an ensemble of seven Efficientnet-B7 models trained with strong augmentation and a frame-by-frame analysis.

\vspace{2mm}
\noindent
\bb{FTCN (Fully Temporal Convolution Network)~} \cite{Zheng2021exploring}, it focuses on temporal cues, exploiting short-term flickering with a Fully Temporal Convolution Network and long-term incoherence with a Temporal Transformer.

\vspace{2mm}
\noindent
\bb{LipForensics~~} \cite{Haliassos2021lips} uses a spatio-temporal network, pre-trained to perform visual speech recognition (lipreading), to detect semantic irregularities in the mouth movements.

\vspace{2mm}
\noindent
\bb{Real Forensics \cite{haliassos2022leveraging}} is a teacher-student network that uses the audio-video pair in a multitask fashion. The student network performs both real/fake classification and features extraction. The teacher is used to provide the target features.

\vspace{2mm}
\noindent
\bb{MDS-based~FD~} \cite{Chugh2020notmade} is an audio-visual fake detector (FD) based on modality dissonance score (MDS), a similarity measure between audio and visual streams. The idea is to capture inconsistencies such
as lack of lip synchronization, unnatural facial movements or asymmetries.

\vspace{2mm}
\noindent
\bb{Joint AV} \cite{Zhou2021joint} is a multimodal detector that handles video and audio streams separately with their own labels. On top of this, the model also synchronizes the representations from the two streams to discriminate synchronization patterns between pristine and manipulated data.

\vspace{2mm}
\noindent
\bb{ICT (Identity Consistency Transformer)~} \cite{dong2022protecting}. It focuses on finding identity-based inconsistencies between the inner region and outer one of the face using a transformer-based architecture. A reference-assisted variant (\bb{ICT-Ref}) is also proposed, that assumes the availability of a reference set of real videos.
    
\vspace{2mm}
\noindent
\bb{ID-Reveal~} \cite{Cozzolino2021idreveal} is a single-modality (visual information only) identity-based detector. It relies on 3D Morphable Models and an adversarial game to improve the discrimination performance.

\section{Deepfake video datasets}
\label{datasets}

In this Section we briefly describe the four deepfake video datasets used in our experiments:

\vspace{2mm}
\noindent
\bb{pDFDC}, preview DeepFake Detection Challenge dataset \cite{Dolhansky2019preview}. We show results on $44$ individuals which have more than $9$ videos with a total of $920$ real videos and $2925$ fake ones.

\vspace{2mm}
\noindent
\bb{FakeAVCelebV2}, Audio-Video Deepfake dataset ~\cite{Khalid2021fakeavceleb}.
It comprises 500 real videos coming from Voxceleb2 \cite{Chung2018voxceleb2} and about 20,000 fake videos
generated by both face-swapping (Faceswap~\cite{Korshunova2017fast}, Faceswap GAN (FSGAN)~\cite{Nirkin2019fsgan})
and facial reenactment (Wav2Lip~\cite{Prajwal2020lip}) 
methods.
Fake audios are generated by a transfer learning-based real-time voice cloning tool (SV2TTS \cite{Jia2018transfer}).
These methods are then used individually or combined together giving rise to five categories of manipulated videos.

\vspace{2mm}
\noindent
\bb{KoDF}, a large-scale Korean DeepFake dataset \cite{Kwon2021kodf}.
It includs three face swapping manipulations: FaceSwap~\cite{FaceSwap2020}, DeepFaceLab~\cite{Petrov2020deep} and FSGAN~\cite{Nirkin2019fsgan}
and three face-reenactment ones:
First Order Motion Model (FOMM)~\cite{Siarohin2019first}, Audio-driven face synthesis ATFHP~\cite{Yi2020audio} and Wav2Lip~\cite{Prajwal2020lip}.
We consider a test-set comprising 276 real videos and 544 fake ones.

\vspace{2mm}
\noindent
\bb{DF-TIMIT}, DeepFake-TIMIT \cite{Korshunov2018deepfakes}.
An open source GAN-based face swapping method is used \cite{FaceSwapGAN}
with two different input dimensions of the GAN network so as to obtain two manipulated videos for each real one.
We report results for videos of at least $4$ seconds, for a total of $290$ real videos and $580$ fake ones.

\section{Limitations}
\label{limitations}
\begin{figure}[t]
    \centering
    \includegraphics[page=6, width=1.\linewidth, trim=0 165 0 0]{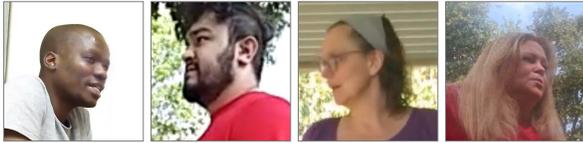}
    \caption{Examples of videos on which our approach fails. All videos come from the pDFDC dataset \cite{Dolhansky2019preview}.}
    \label{fig:errors}
\end{figure}

While our method does not need to include any fake videos in training, 
it requires a large dataset of audio-visual information from several different subjects for training 
and it needs a few (around 10) reference pristine videos of the target subject at testing time. 
We also noticed a drop in performance on faces seen in profile (Fig.\ref{fig:errors}). 
This is probably due to the preponderance of frontal-pose videos in the training set 
and could probably be solved by a better balancing of training samples or by suitable forms of augmentation.

\end{appendix}

{\small
\bibliographystyle{ieee_fullname}
\bibliography{egbib}

\begin{thebibliography}{10}\itemsep=-1pt

\bibitem{FaceSwap2020}
Faceswap.
\newblock \url{https://github.com/deepfakes/faceswap}.

\bibitem{FaceSwapGAN}
{FaceSwap-GAN}.
\newblock \url{https://github.com/shaoanlu/faceswap-GAN}.

\bibitem{Agarwal2020}
S. Agarwal, H. Farid, T. El-Gaaly, and S. Lim.
\newblock Detecting deep-fake videos from appearance and behavior.
\newblock In {\em IEEE international workshop on information forensics and
  security (WIFS)}, 2020.

\bibitem{Agarwal2020detecting}
S. Agarwal, H. Farid, O. Fried, and M. Agrawala.
\newblock Detecting deep-fake videos from phoneme-viseme mismatches.
\newblock In {\em IEEE Conference on Computer Vision and Pattern Recognition
  (CVPR) Workshops}, 2020.

\bibitem{Agarwal2019}
S. Agarwal, H. Farid, Y. Gu, M. He, K. Nagano, and H. Li.
\newblock Protecting world leaders against deep fakes.
\newblock In {\em IEEE Conference on Computer Vision and Pattern Recognition
  (CVPR) Workshops}, 2019.

\bibitem{Agarwal2021watch}
S. Agarwal, L. Hu, E. Ng, T. Darrell, H. Li, and A. Rohrbach.
\newblock Watch those words: Video falsification detection using
  word-conditioned facial motion.
\newblock In {\em IEEE Winter conference on Applications of Computer Vision
  (WACV)}, 2023.

\bibitem{bohavcek2022protecting}
M. Boh{\'a}{\v{c}}ek and H. Farid.
\newblock Protecting world leaders against deep fakes using facial, gestural,
  and vocal mannerisms.
\newblock In {\em Proceedings of the National Academy of Sciences}, 2022.

\bibitem{Bonettini2020video}
N. Bonettini, E.D. Cannas, S. Mandelli, L. Bondi, P. Bestagini, and S. Tubaro.
\newblock {Video Face Manipulation Detection Through Ensemble of CNNs}.
\newblock In {\em IEEE International Conference on Pattern Recognition (ICPR)},
  2020.

\bibitem{Chugh2020notmade}
K. Chugh, P. Gupta, A. Dhall, and R. Subramanian.
\newblock Not made for each other- audio-visual dissonance-based deepfake
  detection and localization.
\newblock In {\em ACM International Conference on Multimedia}, 2020.

\bibitem{Chung2018voxceleb2}
J.S. Chung, A. Nagrani, and A. Zisserman.
\newblock {VoxCeleb2: Deep speaker recognition}.
\newblock In {\em Interspeech}, 2018.

\bibitem{Cozzolino2019extracting}
D. Cozzolino, G. Poggi, and L. Verdoliva.
\newblock Extracting camera-based fingerprints for video forensics.
\newblock In {\em IEEE Conference on Computer Vision and Pattern Recognition
  (CVPR) Workshops}, 2019.

\bibitem{Cozzolino2021idreveal}
D. Cozzolino, A. R{\"o}ssler, J. Thies, M. Nie{\ss}ner, and L. Verdoliva.
\newblock {ID-Reveal: Identity-aware DeepFake Video Detection}.
\newblock In {\em IEEE International Conference on Computer Vision (ICCV)},
  2021.

\bibitem{Cozzolino2021spoc}
D. Cozzolino, J. Thies, A. R{\"{o}}ssler, M. Nie{\ss}ner, and L. Verdoliva.
\newblock {SpoC: Spoofing camera fingerprints}.
\newblock In {\em IEEE Conference on Computer Vision and Pattern Recognition
  (CVPR) Workshops}, 2021.

\bibitem{Cozzolino2018}
D. Cozzolino, J. Thies, A. R{\"{o}}ssler, C. Riess, M. Nie{\ss}ner, and L.
  Verdoliva.
\newblock {ForensicTransfer}: Weakly-supervised domain adaptation for forgery
  detection.
\newblock {\em arXiv preprint arXiv:1812.02510}, 2018.

\bibitem{Dang2020detecting}
H. Dang, F. Liu, J. Stehouwer, X. Liu, and A.K. Jain.
\newblock On the detection of digital face manipulation.
\newblock In {\em IEEE Conference on Computer Vision and Pattern Recognition
  (CVPR)}, 2020.

\bibitem{Das2021towards}
S. Das, S. Seferbekov, A. Datta, Md.~S. Islam, and Md.~R. Amin.
\newblock Towards solving the deepfake problem: An analysis on improving
  deepfake detection using dynamic face augmentation.
\newblock In {\em IEEE International Conference on Computer Vision (ICCV)
  Workshops}, 2021.

\bibitem{Van2008visualizing}
L.~Van der Maaten and G. Hinton.
\newblock {Visualizing data using t-SNE.}
\newblock {\em Journal of machine learning research}, 9(11), 2008.

\bibitem{Dolhansky2020dfdc}
B. Dolhansky, J. Bitton, B. Pflaum, J. Lu, R. Howes, M. Wang, and C.~Canton
  Ferrer.
\newblock The deepfake detection challenge dataset.
\newblock {\em arXiv preprint arXiv:2006.07397}, 2020.

\bibitem{Dolhansky2019preview}
B. Dolhansky, R. Howes, B. Pflaum, N. Baram, and C.~Canton Ferrer.
\newblock {The deepfake detection challenge (DFDC) preview dataset}.
\newblock {\em arXiv preprint arXiv:1910.08854}, 2019.

\bibitem{dong2022protecting}
X. Dong, J. Bao, D. Chen, T. Zhang, W. Zhang, N. Yu, D. Chen, F. Wen, and B.
  Guo.
\newblock Protecting celebrities from deepfake with identity consistency
  transformer.
\newblock In {\em IEEE Conference on Computer Vision and Pattern Recognition
  (CVPR)}, 2022.

\bibitem{Goodfellow2014explaining}
I. Goodfellow, J. Shlens, and C. Szegedy.
\newblock Explaining and harnessing adversarial examples.
\newblock In {\em International Conference on Learning Representations (ICLR)},
  2015.

\bibitem{guo2016celeb}
Y. Guo, L. Zhang, Y. Hu, X. He, and J. Gao.
\newblock {MS-Celeb-1M: A Dataset and Benchmark for Large-Scale Face
  Recognition}.
\newblock In {\em European Conference on Computer Vision (ECCV)}, 2016.

\bibitem{haliassos2022leveraging}
A. Haliassos, R. Mira, S. Petridis, and M. Pantic.
\newblock Leveraging real talking faces via self-supervision for robust forgery
  detection.
\newblock In {\em IEEE Conference on Computer Vision and Pattern Recognition
  (CVPR)}, 2022.

\bibitem{Haliassos2021lips}
A. Haliassos, K. Vougioukas, S. Petridis, and M. Pantic.
\newblock Lips don’t lie: A generalisable and robust approach to face forgery
  detection.
\newblock In {\em IEEE Conference on Computer Vision and Pattern Recognition
  (CVPR)}, 2021.

\bibitem{Khalid2021fakeavceleb}
K. Hasam, T. Shahroz, K. Minha, and S.S. Woo.
\newblock Fake{AVC}eleb: A novel audio-video multimodal deepfake dataset.
\newblock In {\em Thirty-fifth Conference on Neural Information Processing
  Systems Datasets and Benchmarks Track (Round 2)}, 2021.

\bibitem{Heo2020clova}
H.S. Heo, B.-J. Lee, J. Huh, and J.S. Chung.
\newblock Clova baseline system for the voxceleb speaker recognition challenge
  2020.
\newblock {\em arXiv preprint arXiv:2009.14153}, 2020.

\bibitem{Hussain2021adversarial}
S. Hussain, P. Neekhara, M. Jere, F. Koushanfar, and J. McAuley.
\newblock Adversarial deepfakes: Evaluating vulnerability of deepfake detectors
  to adversarial examples.
\newblock In {\em IEEE Winter conference on Applications of Computer Vision
  (WACV)}, 2021.

\bibitem{Jeon2020TGD}
H. Jeon, Y. Bang, J. Kim, and S. Woo.
\newblock {T-GD: Transferable GAN-generated Images Detection Framework}.
\newblock In {\em International Conference on Machine Learning (ICML)}, 2020.

\bibitem{Jia2018transfer}
Y. Jia, Y. Zhang, R. Weiss, Q. Wang, J. Shen, F. Ren, P. Nguyen, R. Pang,
  I.~Lopez Moreno, Y. Wu, et~al.
\newblock Transfer learning from speaker verification to multispeaker
  text-to-speech synthesis.
\newblock In {\em Advances in Neural Information Processing Systems (NeurIPS)},
  2018.

\bibitem{Khan2021video}
S.A. Khan and H. Dai.
\newblock Video transformer for deepfake detection with incremental learning.
\newblock In {\em ACM International Conference on Multimedia}, 2021.

\bibitem{Khosla2020supervised}
P. Khosla, P. Teterwak, C. Wang, A. Sarna, Y. Tian, P. Isola, A. Maschinot, C.
  Liu, and D. Krishnan.
\newblock Supervised contrastive learning.
\newblock In {\em Advances in Neural Information Processing Systems (NeurIPS)},
  2020.

\bibitem{Korshunov2019tampered}
P. Korshunov, M. Halstead, D. Castan, M. Graciarena, M. McLaren, B. Burns, A.
  Lawson, and S. Marcel.
\newblock Tampered speaker inconsistency detection with phonetically aware
  audio-visual features.
\newblock In {\em International Conference on Machine Learning (ICML)
  Workshops}, 2019.

\bibitem{Korshunov2018deepfakes}
P. Korshunov and S. Marcel.
\newblock Deepfakes: a new threat to face recognition? assessment and
  detection.
\newblock {\em arXiv preprint arXiv:1812.08685}, 2018.

\bibitem{Korshunov2018speaker}
P. Korshunov and S. Marcel.
\newblock Speaker inconsistency detection in tampered video.
\newblock In {\em European Signal Processing Conference (EUSIPCO)}, 2018.

\bibitem{Korshunova2017fast}
I. Korshunova, W. Shi, J. Dambre, and L. Theis.
\newblock Fast face-swap using convolutional neural networks.
\newblock In {\em IEEE International Conference on Computer Vision (ICCV)},
  2017.

\bibitem{Kwon2021kodf}
P. Kwon, J. You, G. Nam, S. Park, and G. Chae.
\newblock {KoDF: A large-scale korean deepfake detection dataset}.
\newblock In {\em IEEE International Conference on Computer Vision (ICCV)},
  2021.

\bibitem{Li2021frequency}
J. Li, H. Xie, J. Li, Z. Wang, and Y. Zhang.
\newblock Frequency-aware discriminative feature learning supervised by
  single-center loss for face forgery detection.
\newblock In {\em IEEE Conference on Computer Vision and Pattern Recognition
  (CVPR)}, 2021.

\bibitem{Li2020face}
L. Li, J. Bao, T. Zhang, H. Yang, D. Chen, F. Wen, and B. Guo.
\newblock {Face X-ray for more general face forgery detection}.
\newblock In {\em IEEE Conference on Computer Vision and Pattern Recognition
  (CVPR)}, 2020.

\bibitem{Liu2021spatial}
H. Liu, X. Li, W. Zhou, Y. Chen, Y. He, H. Xue, W. Zhang, and N. Yu.
\newblock Spatial-phase shallow learning: Rethinking face forgery detection in
  frequency domain.
\newblock In {\em IEEE Conference on Computer Vision and Pattern Recognition
  (CVPR)}, 2021.

\bibitem{Loshchilov2017decoupled}
I. Loshchilov and F. Hutter.
\newblock Decoupled weight decay regularization.
\newblock {\em arXiv preprint arXiv:1711.05101}, 2017.

\bibitem{Luo2021generalizing}
Y. Luo, Y. Zhang, J. Yan, and W. Liu.
\newblock Generalizing face forgery detection with high-frequency features.
\newblock In {\em IEEE Conference on Computer Vision and Pattern Recognition
  (CVPR)}, 2021.

\bibitem{Marra2019incremental}
F. Marra, C. Saltori, G. Boato, and L. Verdoliva.
\newblock {Incremental learning for GAN-generated image detection}.
\newblock In {\em IEEE international workshop on information forensics and
  security (WIFS)}, 2019.

\bibitem{Mittal2020}
T. Mittal, U. Bhattacharya, R. Chandra, A. Bera, and D. Manocha.
\newblock Emotions don’t lie: An audio-visual deepfake detection method using
  affective cues.
\newblock In {\em ACM International Conference on Multimedia}, 2018.

\bibitem{Nagrani2020disentangled}
A. Nagrani, J.S. Chung, S. Albanie, and A. Zisserman.
\newblock Disentangled speech embeddings using cross-modal self-supervision.
\newblock In {\em IEEE International Conference on Acoustics, Speech and Signal
  Processing (ICASSP)}, 2020.

\bibitem{Neekhara2021adversarial}
P. Neekhara, B. Dolhansky, J. Bitton, and Cristian~Canton Ferrer.
\newblock Adversarial threats to deepfake detection: A practical perspective.
\newblock In {\em IEEE Conference on Computer Vision and Pattern Recognition
  (CVPR)}, 2021.

\bibitem{Nirkin2019fsgan}
Y. Nirkin, Y. Keller, and T. Hassner.
\newblock {FSGAN: Subject agnostic face swapping and reenactment}.
\newblock In {\em IEEE International Conference on Computer Vision (ICCV)},
  2019.

\bibitem{Petrov2020deep}
I. Petrov, D. Gao, N. Chervoniy, K. Liu, S. Marangonda, C. Um'e, Mr. dpfks, RP
  Luis, J. Jiang, S. Zhang, P. Wu, B. Zhou, and W. Zhang.
\newblock {DeepFaceLab: A simple, flexible and extensible face swapping
  framework}.
\newblock {\em arXiv preprint arXiv:2005.05535}, 2020.

\bibitem{Prajwal2020lip}
K.R. Prajwal, R. Mukhopadhyay, V.P. Namboodiri, and C.V. Jawahar.
\newblock A lip sync expert is all you need for speech to lip generation in the
  wild.
\newblock In {\em ACM International Conference on Multimedia}, 2020.

\bibitem{Roessler2019ff++}
A. R{\"{o}}ssler, D. Cozzolino, L. Verdoliva, C. Riess, J. Thies, and M.
  Nie{\ss}ner.
\newblock Faceforensics++: Learning to detect manipulated facial images.
\newblock In {\em IEEE International Conference on Computer Vision (ICCV)},
  2019.

\bibitem{Sanderson2009multi}
C. Sanderson and B.C. Lovell.
\newblock Multi-region probabilistic histograms for robust and scalable
  identity inference.
\newblock In {\em International Conference on Biometrics (ICB)}, 2009.

\bibitem{Sari2021multi}
L. Sari, K. Singh, J. Zhou, L. Torresani, N. Singhal, and Y. Saraf.
\newblock {A Multi-view approach to audio-visual speaker verification}.
\newblock In {\em IEEE International Conference on Acoustics, Speech and Signal
  Processing (ICASSP)}, 2021.

\bibitem{Selimsef2020}
S. Seferbekov.
\newblock {\em DeepFake Detection (DFDC) Team Sefer.}
\newblock \url{https://github.com/selimsef/dfdc\_deepfake\_challenge}.

\bibitem{Shon2019noise}
S. Shon, T.-H. Oh, and J. Glass.
\newblock Noise-tolerant audio-visual online person verification using an
  attention-based neural network fusion.
\newblock In {\em IEEE International Conference on Acoustics, Speech and Signal
  Processing (ICASSP)}, 2019.

\bibitem{Siarohin2019first}
A. Siarohin, S. Lathuili\`{e}re, S. Tulyakov, E. Ricci, and N. Sebe.
\newblock First order motion model for image animation.
\newblock In {\em Advances in Neural Information Processing Systems (NeurIPS)},
  2019.

\bibitem{Sohn2016improved}
K. Sohn.
\newblock Improved deep metric learning with multi-class n-pair loss objective.
\newblock In {\em Advances in Neural Information Processing Systems (NeurIPS)},
  2016.

\bibitem{Tan2019efficientnet}
M. Tan and Q. Le.
\newblock Efficientnet: Rethinking model scaling for convolutional neural
  networks.
\newblock In {\em International Conference on Machine Learning (ICML)}, 2019.

\bibitem{Tolosana2020DeepFakes}
R. Tolosana, R. Vera-Rodriguez, J. Fierrez, A. Morales, and J. Ortega-Garcia.
\newblock Deepfakes and beyond: A survey of face manipulation and fake
  detection.
\newblock {\em Information Fusion}, pages 131--148, 2020.

\bibitem{Verdoliva2020media}
L. Verdoliva.
\newblock Media forensics and deepfakes: an overview.
\newblock {\em IEEE Journal of Selected Topics in Signal Processing},
  14(5):910--932, 2020.

\bibitem{Wang2021representative}
C. Wang and W. Deng.
\newblock Representative forgery mining for fake face detection.
\newblock In {\em IEEE Conference on Computer Vision and Pattern Recognition
  (CVPR)}, 2021.

\bibitem{Wang2021M2TR}
J. Wang, Z. Wu, X. Ouyang, W.and~Han, J. Chen, Y.-G. Jiang, and S.-N. Li.
\newblock M2tr: Multi-modal multi-scale transformers for deepfake detection.
\newblock In {\em International Conference on Multimedia Retrieval}, 2022.

\bibitem{Wu2018group}
Y. Wu and K. He.
\newblock Group normalization.
\newblock In {\em European Conference on Computer Vision (ECCV)}, 2018.

\bibitem{Yi2020audio}
R. Yi, Z. Ye, J. Zhang, H. Bao, and Y.-J. Liu.
\newblock Audio-driven talking face video generation with learning-based
  personalized head pose.
\newblock {\em arXiv preprint arXiv:2002.10137}, 2020.

\bibitem{Zhao2021multiattentional}
H. Zhao, W. Zhou, D. Chen, T. Wei, W. Zhang, and N. Yu.
\newblock Multi-attentional deepfake detection.
\newblock In {\em IEEE Conference on Computer Vision and Pattern Recognition
  (CVPR)}, 2021.

\bibitem{Zhao2021learning}
T. Zhao, X. Xu, M. Xu, H. Ding, Y. Xiong, and W. Xia.
\newblock Learning self-consistency for deepfake detection.
\newblock In {\em IEEE International Conference on Computer Vision (ICCV)},
  2021.

\bibitem{Zheng2021exploring}
Y. Zheng, J. Bao, D. Chen, M. Zeng, and F. Wen.
\newblock Exploring temporal coherence for more general video face forgery
  detection.
\newblock In {\em IEEE International Conference on Computer Vision (ICCV)},
  2021.

\bibitem{Zhou2021joint}
Y. Zhou and S.-N. Lim.
\newblock Joint audio-visual deepfake detection.
\newblock In {\em IEEE International Conference on Computer Vision (ICCV)},
  2021.

\bibitem{Zhu2021face}
X. Zhu, H. Wang, H. Fei, Z. Lei, and S.Z. Li.
\newblock Face forgery detection by 3d decomposition.
\newblock In {\em IEEE Conference on Computer Vision and Pattern Recognition
  (CVPR)}, 2021.

\end{thebibliography}
}

\end{document}